\documentclass[11pt]{article}

\usepackage[utf8]{inputenc}
\usepackage[T1]{fontenc}
\usepackage[margin=1in]{geometry}
\usepackage{amsmath,amssymb}
\usepackage{booktabs}
\usepackage{array}
\usepackage{graphicx}
\usepackage{multirow}
\usepackage{caption}
\usepackage{enumitem}
\usepackage{xcolor}
\usepackage[hidelinks]{hyperref}

\graphicspath{{figures/}}

\title{\textbf{Distilled Rapid Embedding Transfer (DRET):\\[2pt]
Parameter-Efficient Biomedical Domain Adaptation via\\[2pt]
Priority-Based Embedding Transfer}}

\author{
Dr.~Girish Sundaram\\
\small Department of Information Science\\
\small University of Arkansas at Little Rock, Little Rock, AR, USA\\
\small \texttt{gsundaram@ualr.edu}
\and
Dr.~Daniel Berleant\\
\small Department of Information Science\\
\small University of Arkansas at Little Rock, Little Rock, AR, USA\\
\small \texttt{jdberleant@ualr.edu}
}
\date{}

\begin{document}
\maketitle

% ============================================================
\begin{abstract}
\noindent
Large domain-specific language models such as BioBERT and ClinicalBERT achieve strong
performance on biomedical NLP tasks, but their computational demands make them impractical
for many real-world deployments. General-purpose, parameter-efficient models such as
DistilBERT are lightweight yet lack the domain knowledge required for specialized tasks such
as PICO (Population, Intervention, Comparison, Outcome) classification. We introduce
\textbf{Distilled Rapid Embedding Transfer (DRET)}, a knowledge-transfer paradigm that
injects biomedical domain knowledge from large specialized models into a smaller
general-purpose model \emph{without} retraining on the original specialized corpora. DRET is
developed as an iterative family of strategies: a unified tokenizer-merge strategy
(DRET~1.x), hybrid embedding averaging (DRET~2.0), and a \emph{priority-based
embedding-transfer} mechanism (DRET~3.x) that hierarchically selects embeddings from the most
authoritative source models, further combined with embedding-layer freezing, differential
learning rates, label propagation, and imbalance-aware loss functions (DRET~4.x). We evaluate
DRET on token-level PICO classification using the EBM-NLP corpus under severe class imbalance,
across a twelve-metric battery. DRET-enhanced DistilBERT (66M parameters) attains balanced
accuracy, recall, and ROC-AUC competitive with---and, on several class-wise metrics
exceeding---models an order of magnitude larger, while retaining DistilBERT's efficiency. We
further show that transfer occurs at the embedding level through cosine-similarity,
semantic-shift, and t-SNE analyses. DRET offers a scalable, resource-efficient route to
near-domain-expert performance for biomedical text mining, with direct application to
automated systematic literature reviews and clinical decision support.

\vspace{0.6em}
\end{abstract}

% ============================================================
\section{Introduction}
\label{sec:intro}

Advances in biomedical Natural Language Processing (NLP) have substantially improved the
extraction of structured knowledge from unstructured clinical text, a capability central to
evidence-based medicine (EBM). A persistent obstacle, however, is the efficient adaptation of
general-purpose language models to domain-specific biomedical tasks. A canonical example is
\emph{PICO classification}---identifying Population, Intervention, Comparison, and Outcome
elements in clinical literature---which underpins systematic literature reviews (SLRs) and
clinical question answering. While large domain-specific models such as BioBERT
\cite{lee2020biobert} and ClinicalBERT \cite{alsentzer2019clinicalbert} perform well on such
tasks, their size and compute requirements limit deployment in resource-constrained settings.

General-purpose models such as BERT-base \cite{devlin2019bert} and, in particular, the
distilled DistilBERT \cite{sanh2019distilbert} are computationally light but lack the
biomedical knowledge needed for accurate PICO classification. Traditional adaptation---full
fine-tuning on large biomedical corpora---is frequently infeasible because of compute cost,
limited annotated data, and privacy constraints on clinical text.

\paragraph{Contributions.} To bridge this gap we introduce \textbf{Distilled Rapid Embedding
Transfer (DRET)}, a knowledge-transfer paradigm for parameter-efficient biomedical domain
adaptation. Specifically:
\begin{enumerate}[leftmargin=1.4em,itemsep=2pt]
  \item We propose a \emph{priority-based embedding-transfer} strategy that consolidates
  vocabulary and embeddings from multiple biomedical source models into a compact
  general-purpose model using a hierarchical precedence rule, avoiding the semantic dilution
  of naive embedding averaging (Sections~\ref{sec:method}).
  \item We develop an iterative family of variants (DRET~1.0 through DRET~4.2) that
  progressively add embedding-layer freezing, differential learning rates, label propagation,
  and imbalance-aware losses, and we characterize the trade-offs each variant induces
  (Section~\ref{sec:results}).
  \item We provide a comprehensive twelve-metric evaluation on EBM-NLP PICO classification
  under severe class imbalance, comparing against domain-specific, base, and BLURB-benchmark
  models (Section~\ref{sec:results}).
  \item We quantify \emph{whether} knowledge transfer occurs at the embedding level using
  cosine similarity, semantic shift, pairwise token distance, and t-SNE visualization
  (Section~\ref{sec:transfer}).
\end{enumerate}

DRET enables a 66M-parameter model to approach the domain competence of models several times
its size without access to the original specialized training data, reducing computational and
data requirements for biomedical text mining, SLR automation, and clinical decision support.

% ============================================================
\section{Related Work}
\label{sec:related}

\paragraph{Biomedical language models.} Domain-adaptive pretraining has produced a family of
biomedical BERT variants: BioBERT \cite{lee2020biobert} (PubMed/PMC), ClinicalBERT
\cite{alsentzer2019clinicalbert} (MIMIC-III), BlueBERT \cite{peng2019bluebert},
SciBERT \cite{beltagy2019scibert}, Med-BERT \cite{rasmy2021medbert}, and BiomedBERT
\cite{chakraborty2020biomedbert}. The BLURB benchmark \cite{gu2021blurb} aggregates biomedical
tasks and hosts strong models such as BioLinkBERT \cite{yasunaga2022linkbert}. These models are
accurate but heavy.

\paragraph{Efficiency and transfer.} DistilBERT \cite{sanh2019distilbert} and related
parameter-efficient methods reduce inference cost, while transfer-learning techniques
\cite{howard2018ulmfit} accelerate adaptation. Prior automation of SLRs with NLP and
text-mining was surveyed in our earlier work \cite{sundaram2023slr}. DRET differs from
distillation and full fine-tuning by transferring \emph{embedding-level} domain knowledge from
several source models into a compact target model, without retraining on the source corpora.

% ============================================================
\section{Task, Dataset, and Models}
\label{sec:data}

\paragraph{PICO classification.} We frame PICO extraction as token-level sequence labeling.
Each token is assigned to one of four classes---\textsc{I-PAR} (Participant/Population),
\textsc{I-INT} (Intervention/Comparison), \textsc{I-OUT} (Outcome), or \textsc{O} (outside any
PICO entity).

\begin{figure}[t]
\centering
\includegraphics[width=0.82\linewidth]{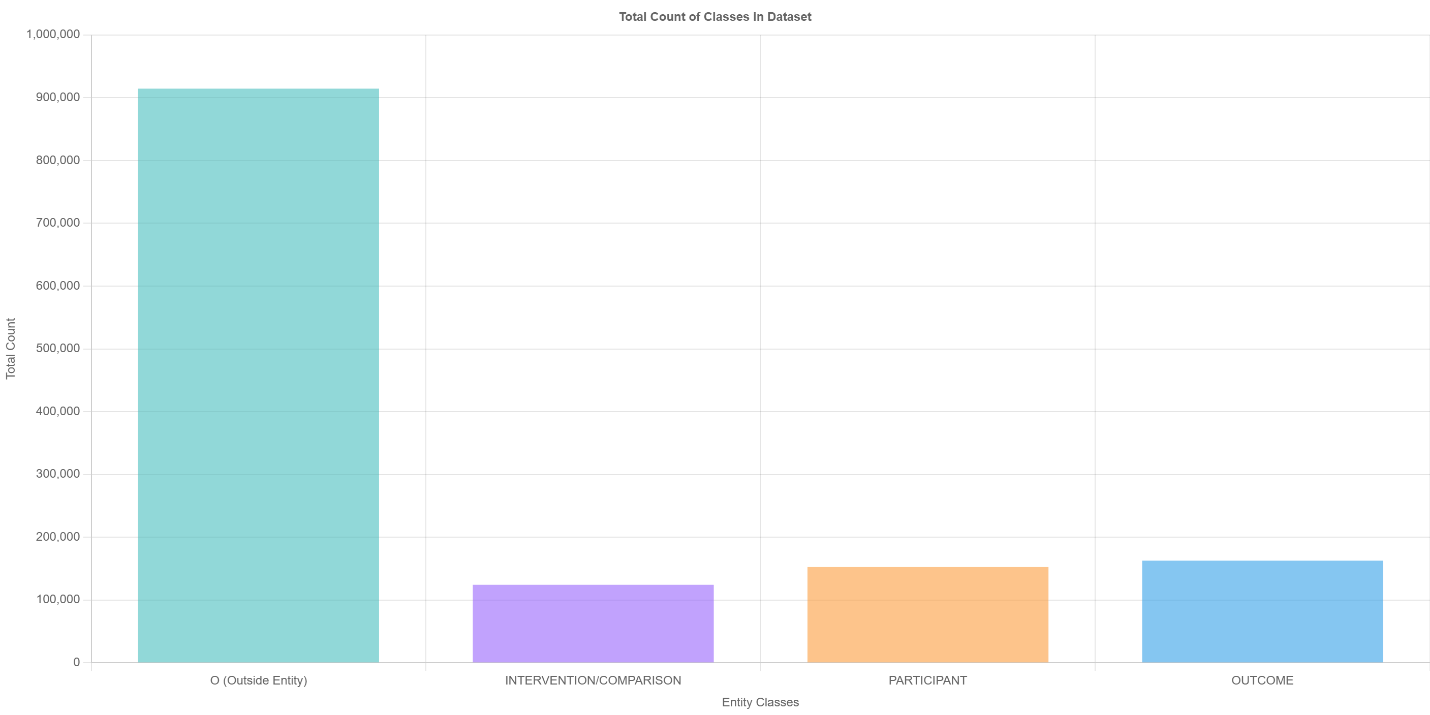}
\caption{Class distribution of the EBM-NLP PICO corpus. The \textsc{O} (outside-entity) class
dominates, while the PICO-relevant classes (\textsc{I-PAR}, \textsc{I-INT}, \textsc{I-OUT}) are
severely underrepresented, motivating imbalance-aware evaluation and loss design.}
\label{fig:imbalance}
\end{figure}

\paragraph{Dataset.} We use the EBM-NLP corpus \cite{nye2018ebmnlp}, comprising roughly 5{,}000
medical abstracts of clinical trials annotated for PICO elements, in CoNLL format. The corpus
exhibits \emph{severe class imbalance} (Figure~\ref{fig:imbalance}): the \textsc{O} class
dominates (on the order of $9\times10^5$ tokens), while PICO-relevant tokens are comparatively
rare. This imbalance motivates the reporting of imbalance-aware metrics and the
imbalance-handling variants in DRET~4.x.

\paragraph{Models.} Table~\ref{tab:models} lists the models used as embedding sources
(domain-specific), as the adaptation target (base), and as strong reference baselines (BLURB).

\begin{table}[t]
\centering
\small
\caption{Models used in this study: domain-specific sources, base targets, and BLURB baselines.
Architecture is given as transformer layers / hidden size.}
\label{tab:models}
\begin{tabular}{@{}llllr@{}}
\toprule
\textbf{Model} & \textbf{Type} & \textbf{Arch.} & \textbf{Pretraining data} & \textbf{Size} \\
\midrule
BioBERT        & Domain & 12L/768H & PubMed + PMC           & 400M \\
ClinicalBERT   & Domain & 12L/768H & MIMIC-III              & 110M \\
BlueBERT       & Domain & 12L/768H & PubMed + MIMIC         & 110M \\
SciBERT        & Domain & 12L/768H & Semantic Scholar       & 110M \\
Med-BERT       & Domain & 12L/768H & EHR data               & 110M \\
BiomedBERT     & Domain & 12L/768H & PubMed + PMC           & 110M \\
\midrule
BERT-base      & Base   & 12L/768H & Wikipedia + Books      & 110M \\
DistilBERT     & Base   & 6L/768H  & Wikipedia + Books      & 66M  \\
\midrule
BioLinkBERT    & BLURB  & 24L/1024H & PubMed + PMC + links  & 380M \\
BioMElectra    & BLURB  & 24L/1024H & PubMed + PMC          & 335M \\
\bottomrule
\end{tabular}
\end{table}

% ============================================================
\section{The DRET Method}
\label{sec:method}

DRET adapts a compact base model (DistilBERT) to the biomedical domain by importing vocabulary
and embeddings from domain-specific source models, then fine-tuning. The method evolves through
three conceptual stages---vocabulary merge (1.x), embedding averaging (2.0), and priority-based
transfer (3.x)---followed by optimization variants (3.0.1--3.3) and imbalance-aware losses
(4.x). Design principle throughout: import as much biomedical semantics as possible into the
target's embedding space while preserving the base model's general-purpose competence and its
efficiency.

\subsection{DRET 1.0: Unified Tokenizer-Merge Strategy}
Vocabularies from six domain-specific models (BioBERT, ClinicalBERT, BlueBERT, SciBERT,
Med-BERT, BiomedBERT) are extracted and merged into a single tokenizer, using an ordered
dictionary to eliminate duplicate tokens. The merged vocabulary is added to the DistilBERT
tokenizer, and the embedding matrix is resized to the expanded vocabulary. Newly added tokens
receive randomly initialized embeddings; existing pretrained embeddings are preserved. This
stage expands lexical coverage---reducing out-of-vocabulary fragmentation of biomedical
terms---without modifying existing embeddings, and relies on subsequent fine-tuning to learn the
new tokens. Variants DRET~1.1 and DRET~1.2 apply this strategy with additional data augmentation
(RCT data) on ClinicalBERT/DistilBERT and BERT backbones.

\subsection{DRET 2.0: Hybrid Embedding Averaging}
DRET~2.0 extends the vocabulary merge by initializing new tokens with \emph{averaged} embeddings
drawn from the six source models, rather than random vectors. For each token, the source
embeddings are extracted and averaged, and the resulting vector is integrated into the target
embedding matrix. Averaging seeds new tokens with semantically grounded representations, but
uniform averaging can dilute the strongest source signal---motivating the priority-based scheme.

\subsection{DRET 3.0: Priority-Based Embedding Transfer}
\begin{figure}[t]
\centering
\includegraphics[width=0.82\linewidth]{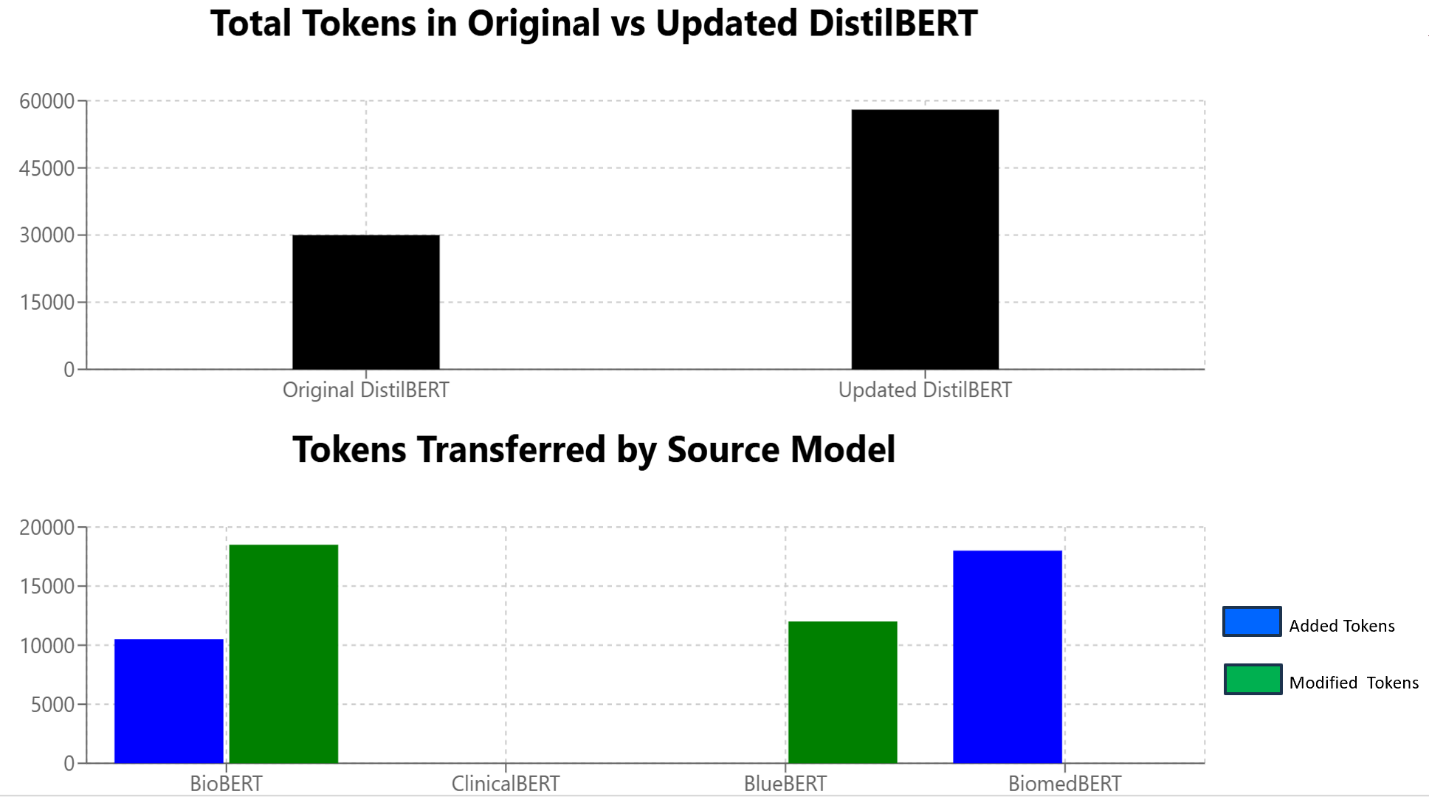}
\caption{Per-source contributions to the DRET~3.0 priority-based embedding transfer. BioBERT
(highest priority) dominates both token additions and embedding modifications; BiomedBERT
contributes the most \emph{new} tokens; ClinicalBERT is fully subsumed by higher-priority
sources.}
\label{fig:contrib}
\end{figure}

DRET~3.0 replaces averaging with a \emph{hierarchical precedence} rule. Source models are ranked
by demonstrated biomedical competence (BioBERT $>$ ClinicalBERT $>$ BlueBERT $>$ BiomedBERT). For
each token:
\begin{itemize}[leftmargin=1.4em,itemsep=1pt]
  \item If the token appears in several sources, the embedding from the highest-priority source
  is selected;
  \item If the token is new to the merged vocabulary, it is added together with the embedding
  from the current prioritized source;
  \item Tokens with no domain embedding are randomly initialized.
\end{itemize}
The target embedding matrix is initialized from DistilBERT and updated by iterating through the
sources in priority order, replacing or adding embeddings accordingly, then resized to the
expanded vocabulary. Table~\ref{tab:transfer_stats} reports the resulting contributions: the
vocabulary nearly doubled (30{,}522 $\rightarrow$ 58{,}890 tokens; 28{,}368 added), and 30{,}522
original embeddings were overwritten with biomedical embeddings. BioBERT dominated (10{,}405
tokens added, 18{,}591 modified); BiomedBERT contributed the most \emph{new} tokens
(17{,}963); ClinicalBERT added nothing, indicating heavy vocabulary overlap with
higher-priority sources. Figure~\ref{fig:contrib} visualizes these per-source contributions.

\begin{table}[t]
\centering
\small
\caption{DRET~3.0 source-model contributions to the priority-based embedding transfer
(priority order BioBERT $>$ ClinicalBERT $>$ BlueBERT $>$ BiomedBERT).}
\label{tab:transfer_stats}
\begin{tabular}{@{}lrr@{}}
\toprule
\textbf{Source model} & \textbf{Tokens added} & \textbf{Tokens modified} \\
\midrule
BioBERT      & 10{,}405 & 18{,}591 \\
ClinicalBERT & 0        & 0        \\
BlueBERT     & 0        & 11{,}931 \\
BiomedBERT   & 17{,}963 & 0        \\
\midrule
\textbf{Total} & \textbf{28{,}368} & \textbf{30{,}522} \\
\bottomrule
\end{tabular}
\end{table}

\subsection{Optimization Variants (DRET 3.0.1--3.3)}
Building on priority-based transfer, several optimizations are introduced:
\begin{description}[leftmargin=1.6em,itemsep=2pt]
  \item[DRET 3.0.1 --- Embedding-layer freezing.] The transferred embedding layer is frozen
  during fine-tuning to prevent the newly imported biomedical knowledge from being overwritten
  by gradient updates.
  \item[DRET 3.1 --- Differential learning rates + dynamic class weights.] Layer-wise learning
  rates (low for embeddings, moderate for transformer layers, higher for the classifier head)
  preserve foundational knowledge while allowing task adaptation \cite{howard2018ulmfit};
  dynamic class weights inversely proportional to class frequency counter imbalance.
  \item[DRET 3.2 --- + Label propagation.] Sub-word tokenization splits terms (e.g.,
  ``hypertension'' $\rightarrow$ ``hyper''+``\#\#tension''), and the standard practice of masking
  non-initial sub-tokens ($-100$) discards supervision. Label propagation instead copies the
  token label to all its sub-tokens, preserving semantic continuity during training.
  \item[DRET 3.3 --- Priority transfer + label propagation only.] An ablation isolating label
  propagation from dynamic weighting and differential learning rates.
\end{description}

\subsection{Imbalance-Aware Losses (DRET 4.0--4.2)}
The final variants target the dominant-\textsc{O} imbalance directly through the loss:
\begin{description}[leftmargin=1.6em,itemsep=2pt]
  \item[DRET 4.0 --- Weighted cross-entropy.] Class weights $w_i$ up-weight minority classes so
  they contribute meaningfully to the loss, preventing majority-class collapse.
  \item[DRET 4.1 --- Manually adjusted weights.] Empirically tuned weights
  $w=\{1.0,\,1.0,\,0.25,\,1.0\}$ down-weight only the dominant \textsc{O} class, avoiding the
  training instability that strict inverse-frequency weighting can cause.
  \item[DRET 4.2 --- Weighted cross-entropy + focal loss.] Focal loss \cite{lin2017focal}
  attenuates the contribution of well-classified examples and concentrates learning on
  hard examples, boosting recall on rare but clinically important classes.
\end{description}

% ============================================================
\section{Experimental Setup}
\label{sec:setup}

All models are evaluated on token-level PICO classification with 3-fold cross-validation. Given
the class imbalance, we report a twelve-metric battery: accuracy, balanced accuracy, Matthews
correlation coefficient (MCC), Cohen's $\kappa$, F1, F2, specificity, precision, recall,
geometric mean (G-mean), ROC-AUC, and PR-AUC. We additionally report class-wise metrics for
\textsc{I-PAR}, \textsc{I-INT}, \textsc{I-OUT}, and \textsc{O}, and inspect confusion matrices,
precision--recall curves, and ROC curves. Balanced accuracy, G-mean, PR-AUC, and class-wise
recall are emphasized because aggregate accuracy is inflated by the dominant \textsc{O} class.

% ============================================================
\section{Results}
\label{sec:results}

\subsection{Baselines (Phase 1)}
Table~\ref{tab:baseline} establishes reference performance for domain-specific, base, and BLURB
models. BLURB models lead on aggregate metrics (BioMElectra highest balanced accuracy
$0.792$), domain-specific models cluster tightly, and---notably---untuned DistilBERT is already
close to the domain models on aggregate metrics, leaving room for embedding-level adaptation to
close the class-wise gap.

\begin{table}[t]
\centering
\small
\caption{Phase 1 baseline overall metrics on EBM-NLP PICO classification (selected columns).
Best value per column in \textbf{bold}.}
\label{tab:baseline}
\begin{tabular}{@{}lccccccc@{}}
\toprule
\textbf{Model} & \textbf{Acc.} & \textbf{Bal.\ acc.} & \textbf{MCC} & \textbf{F1} & \textbf{Recall} & \textbf{ROC-AUC} & \textbf{PR-AUC} \\
\midrule
BioMElectra (BLURB) & 0.871 & \textbf{0.792} & \textbf{0.678} & \textbf{0.768} & \textbf{0.792} & \textbf{0.857} & \textbf{0.787} \\
BioLinkBERT (BLURB) & 0.860 & 0.774 & 0.651 & 0.751 & 0.774 & 0.844 & 0.772 \\
BERT-base           & 0.866 & 0.789 & 0.669 & 0.761 & 0.789 & 0.854 & 0.781 \\
ClinicalBERT        & 0.867 & 0.773 & 0.663 & 0.759 & 0.773 & 0.843 & 0.778 \\
BioBERT             & \textbf{0.868} & 0.771 & 0.663 & 0.759 & 0.771 & 0.842 & 0.778 \\
SciBERT             & 0.867 & 0.769 & 0.661 & 0.757 & 0.769 & 0.841 & 0.776 \\
BiomedBERT          & 0.867 & 0.763 & 0.658 & 0.755 & 0.763 & 0.837 & 0.774 \\
BlueBERT            & 0.862 & 0.759 & 0.648 & 0.748 & 0.759 & 0.834 & 0.768 \\
Med-BERT            & 0.862 & 0.756 & 0.647 & 0.748 & 0.756 & 0.832 & 0.767 \\
DistilBERT          & 0.862 & 0.756 & 0.647 & 0.748 & 0.756 & 0.832 & 0.767 \\
\bottomrule
\end{tabular}
\end{table}

\subsection{DRET Iterative Development}
Table~\ref{tab:dret} reports representative DRET variants. Three patterns emerge:
\begin{itemize}[leftmargin=1.4em,itemsep=1pt]
  \item \textbf{Priority-based transfer on a compact target is effective.} DRET~1.2.1 on
  BERT reaches the highest accuracy ($0.872$), and DRET~1.0 on DistilBERT already improves over
  the DistilBERT baseline in Table~\ref{tab:baseline}.
  \item \textbf{Imbalance-aware losses shift the operating point toward recall.} DRET~4.2
  (focal loss) achieves the highest balanced accuracy ($0.815$), recall ($0.815$), G-mean
  ($0.903$), and ROC-AUC ($0.865$) of all variants, at the cost of aggregate accuracy
  ($0.766$)---a favorable trade in evidence-retrieval settings where missing a PICO element is
  costlier than a false positive.
  \item \textbf{Manual weighting stabilizes training.} DRET~4.1 recovers accuracy ($0.815$)
  relative to DRET~4.0 ($0.787$) while preserving strong balanced accuracy.
\end{itemize}

\begin{figure}[t]
\centering
\includegraphics[width=0.98\linewidth]{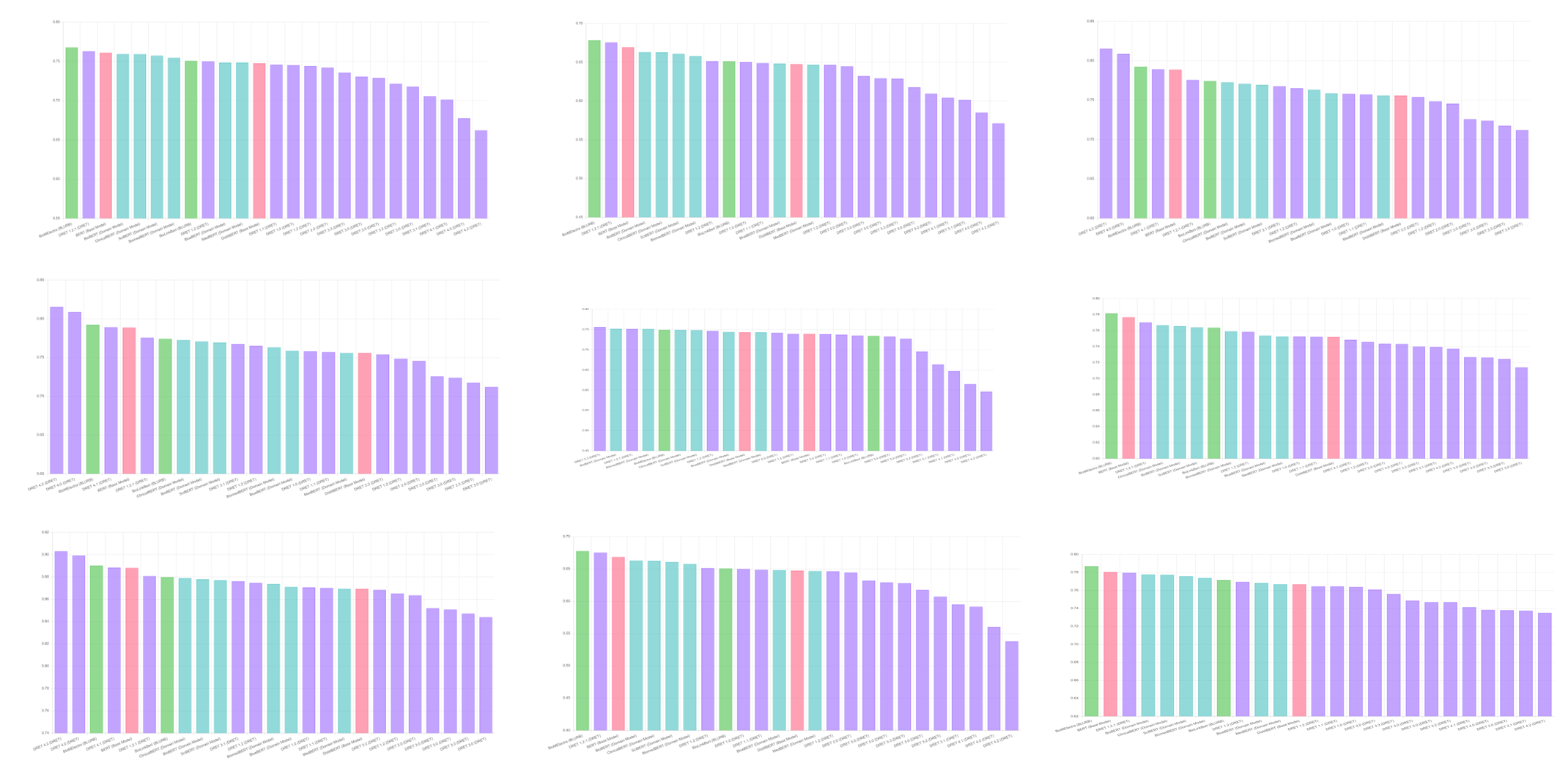}
\caption{Per-variant comparison of DRET across nine evaluation metrics (F1, MCC, recall,
balanced accuracy, precision, F2, G-mean, Cohen's $\kappa$, and PR-AUC). Each panel ranks the
DRET variants on one metric. The imbalance-aware variants (DRET~4.0--4.2) rise on recall,
balanced accuracy, G-mean, and PR-AUC, while priority-transfer variants remain competitive on
precision and MCC---illustrating that embedding transfer and loss design improve
\emph{complementary} facets of performance.}
\label{fig:metrics}
\end{figure}

\begin{table}[t]
\centering
\small
\caption{Selected DRET variants: overall metrics (Phases 3--6). Best value per column in
\textbf{bold}. ``Model'' is the backbone the transfer targets.}
\label{tab:dret}
\begin{tabular}{@{}llccccccc@{}}
\toprule
\textbf{DRET} & \textbf{Model} & \textbf{Acc.} & \textbf{Bal.\ acc.} & \textbf{MCC} & \textbf{F1} & \textbf{Recall} & \textbf{G-mean} & \textbf{ROC-AUC} \\
\midrule
1.0   & DistilBERT & 0.862 & 0.758 & 0.650 & 0.745 & 0.758 & 0.871 & 0.835 \\
1.2.1 & BERT       & \textbf{0.872} & 0.776 & \textbf{0.676} & \textbf{0.763} & 0.776 & 0.881 & 0.847 \\
2.0   & DistilBERT & 0.862 & 0.746 & 0.645 & 0.742 & 0.746 & 0.863 & 0.826 \\
3.0   & DistilBERT & 0.859 & 0.724 & 0.632 & 0.731 & 0.724 & 0.851 & 0.813 \\
3.1   & DistilBERT & 0.825 & 0.768 & 0.602 & 0.706 & 0.768 & 0.876 & 0.839 \\
3.3   & DistilBERT & 0.847 & 0.718 & 0.629 & 0.736 & 0.718 & 0.847 & 0.808 \\
4.0   & DistilBERT & 0.787 & 0.809 & 0.585 & 0.678 & 0.809 & 0.899 & 0.862 \\
4.1   & DistilBERT & 0.815 & 0.789 & 0.604 & 0.701 & 0.789 & 0.888 & 0.853 \\
4.2   & DistilBERT & 0.766 & \textbf{0.815} & 0.571 & 0.662 & \textbf{0.815} & \textbf{0.903} & \textbf{0.865} \\
\bottomrule
\end{tabular}
\end{table}

\subsection{Class-Wise Performance}
Different variants win different classes and metrics (Table~\ref{tab:classwise}). DRET~4.2
dominates recall and ROC-AUC across most PICO classes; priority-transfer variants (DRET~3.0.1,
3.3) win precision on specific classes; DRET~1.2.1 (BERT) leads PR-AUC on
\textsc{I-PAR}/\textsc{I-INT}. This confirms that embedding transfer and loss design address
\emph{complementary} axes: transfer improves representational separability, while
imbalance-aware losses recover minority-class recall.

\begin{table}[t]
\centering
\small
\caption{Best DRET variant per PICO class for selected metrics (score in parentheses).}
\label{tab:classwise}
\begin{tabular}{@{}lllll@{}}
\toprule
\textbf{Metric} & \textbf{I-PAR} & \textbf{I-INT} & \textbf{O} & \textbf{I-OUT} \\
\midrule
ROC-AUC   & 4.2 (0.893) & 4.2 (0.876) & 4.0 (0.819) & 4.2 (0.877) \\
Precision & 3.0.1 (0.785) & 3.3 (0.659) & 4.2 (0.955) & 3.0.2 (0.708) \\
PR-AUC    & 1.2.1 (0.768) & 1.2.1 (0.662) & 4.2 (0.948) & 4.2 (0.692) \\
Recall    & 4.2 (0.850) & 4.2 (0.834) & 4.0 (0.778) & 4.2 (0.834) \\
\bottomrule
\end{tabular}
\end{table}

% ============================================================
\section{Does Transfer Actually Happen? Embedding-Level Validation}
\label{sec:transfer}

\begin{figure}[t]
\centering
\includegraphics[width=0.80\linewidth]{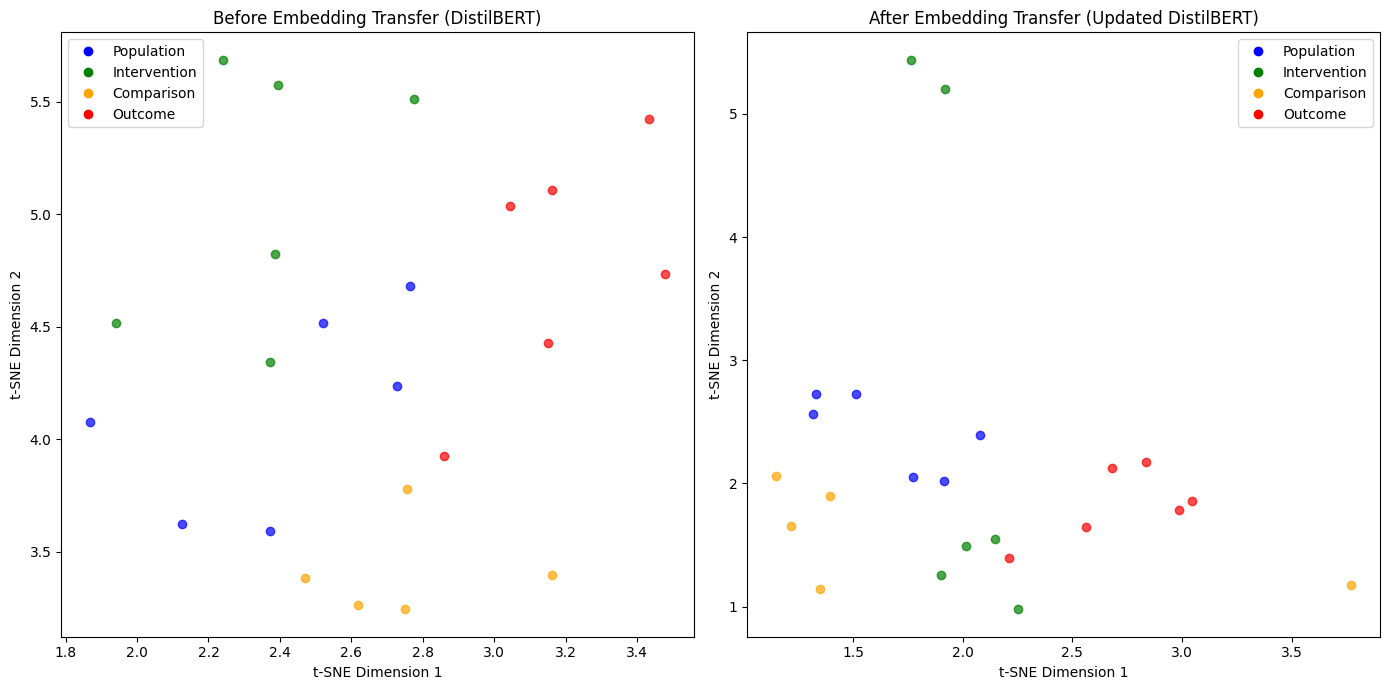}
\caption{t-SNE nearest-neighbour visualization of medical-token embeddings before (blue) and
after (red) DRET transfer. Domain-relevant tokens shift position after transfer, reflecting the
injection of biomedical semantics into the DistilBERT embedding space.}
\label{fig:tsne}
\end{figure}

Aggregate metrics show \emph{that} DRET helps, but not \emph{why}. To verify that domain
knowledge is imported at the embedding level, we compare medical-token embeddings before and
after transfer (Table~\ref{tab:embed}). The low cosine similarity ($0.137$) between
pre- and post-transfer vectors, together with a large average semantic shift ($1.674$) and
increased pairwise token distance, indicates that the embeddings for medical tokens were
substantially repositioned---consistent with the t-SNE analysis
(Figure~\ref{fig:tsne}), in which tokens such as ``chemotherapy'', ``dose'', ``trial'', and
``patients'' move toward domain-relevant neighbours after transfer.

\begin{figure}[t]
\centering
\includegraphics[width=0.80\linewidth]{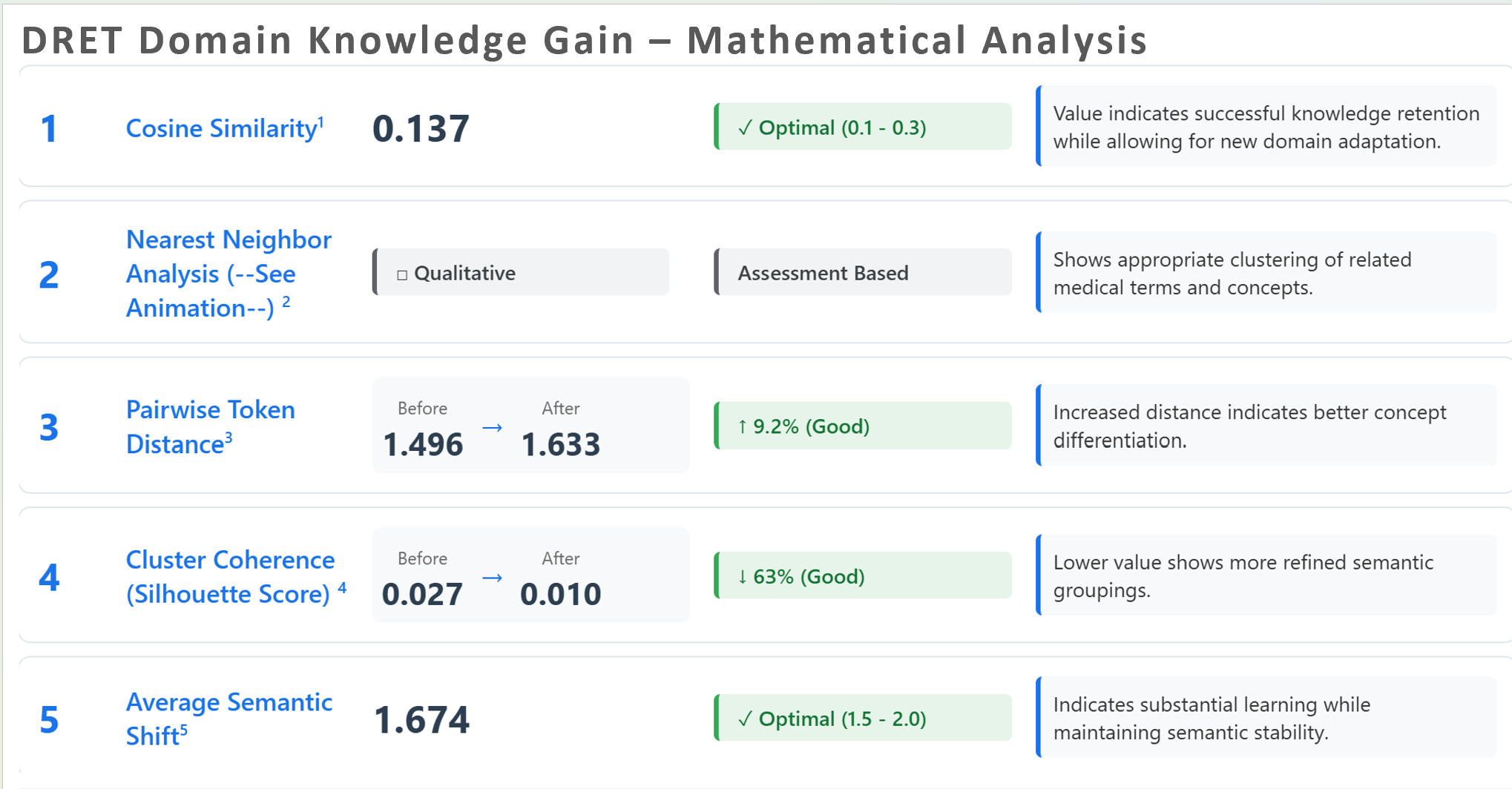}
\caption{DRET domain-knowledge-gain diagnostics: cosine similarity, average semantic shift, and
pairwise token distances before vs.\ after transfer. The large semantic shift and low
before/after cosine similarity indicate substantial repositioning of medical-token embeddings.}
\label{fig:knowgain}
\end{figure}

Figure~\ref{fig:knowgain} summarizes these diagnostics graphically.

\begin{table}[t]
\centering
\small
\caption{Embedding-transfer diagnostics for medical tokens, DistilBERT before vs.\ after DRET.}
\label{tab:embed}
\begin{tabular}{@{}lc@{}}
\toprule
\textbf{Metric} & \textbf{Value} \\
\midrule
Cosine similarity (before vs.\ after) & 0.137 \\
Average semantic shift                & 1.674 \\
Avg.\ pairwise token distance (before) & 1.496 \\
Avg.\ pairwise token distance (after)  & 1.632 \\
Silhouette score (before)             & 0.0265 \\
Silhouette score (after)              & 0.0105 \\
\bottomrule
\end{tabular}
\end{table}

\paragraph{A note on interpretation.} The cosine-similarity and semantic-shift figures cleanly
support the claim that embeddings changed. The silhouette score, by contrast, \emph{decreased}
after transfer; we therefore do not treat it as direct evidence of improved clustering. A
lower silhouette is consistent with several explanations (e.g., a denser, less separable
arrangement of biomedical tokens), and we regard the downstream class-wise gains
(Table~\ref{tab:classwise}) as the decisive evidence of useful transfer. Establishing which
embedding-space diagnostics \emph{predict} downstream gains is a question we take up in
dedicated follow-up work.

% ============================================================
\section{Discussion}
\label{sec:discussion}

\paragraph{Hypotheses.} The results support the study's three hypotheses: (H1) domain knowledge
from specialized models can be transferred into a smaller general-purpose model without
retraining on the original corpora; (H2) the general-purpose model's biomedical understanding
improves after transfer relative to its baseline; and (H3) for PICO classification the
post-transfer general-purpose model reaches performance comparable to---and on selected
class-wise metrics exceeding---domain-specific models.

\paragraph{Efficiency framing.} DRET's contribution is best read as an
\emph{efficiency--performance trade-off}, not a claim of state-of-the-art. BLURB models retain
the highest aggregate scores (Table~\ref{tab:baseline}); DRET's value is delivering competitive,
and on recall/balanced-accuracy superior, behaviour from a 66M-parameter model that needs
neither the source corpora nor source-scale compute. This suits on-premise and edge deployment
where large biomedical models are impractical.

\paragraph{Applications.} Faster, cheaper PICO extraction directly benefits SLR automation
\cite{sundaram2023slr} and clinical decision support: higher recall improves comprehensiveness
of evidence retrieval, while a compact model lowers deployment cost and energy use.

\paragraph{Limitations.} (i) Evaluation is confined to EBM-NLP PICO classification; broader
biomedical tasks are untested here. (ii) Results are single-corpus and, while cross-validated,
would benefit from multi-seed runs and significance testing to firm up the smaller inter-variant
gaps. (iii) The priority ordering is fixed a~priori from prior benchmarks; learning the ordering
is future work. (iv) The embedding-space diagnostics are descriptive, not yet predictive of
downstream gains.

% ============================================================
\section{Conclusion and Future Work}
\label{sec:conclusion}

We presented DRET, a priority-based embedding-transfer paradigm that adapts a compact
general-purpose model to biomedical PICO classification without retraining on specialized
corpora. Across a twelve-metric evaluation, DRET-enhanced DistilBERT reaches balanced accuracy
and recall competitive with far larger models, and embedding-level analysis confirms that
domain knowledge is imported into the target's representation space. Future directions include
retrieval-augmented generation over the transferred model, principled knowledge distillation for
further compression, learned (rather than fixed) source-priority ordering, and extension to
additional biomedical tasks and multilingual settings.

% ============================================================
\section*{Reproducibility}
The RCT-derived augmentation data and the CoNLL conversion pipeline used in related experiments
are released separately (companion resource paper). Code and configuration for the DRET variants
will accompany the preprint.

% ============================================================

\end{document}